\documentclass[letterpaper, 10 pt, conference]{ieeeconf}

\IEEEoverridecommandlockouts                              

\usepackage{cite}
\usepackage{listings}
\usepackage{float}
\usepackage[dvipdfmx]{graphicx}
\usepackage{amssymb}
\usepackage{latexsym}
\usepackage{amsfonts}
\usepackage{url}
\usepackage{comment}
\usepackage{algorithm}
\usepackage{algpseudocode}
\usepackage{amsmath}
\usepackage{listings,jvlisting}
\usepackage{tikz}
\usetikzlibrary{shapes.geometric, arrows}

\lstdefinelanguage{json}{
    morestring=[b]",
    morestring=[s]{'}{'},
    morecomment=[l]{//},
    morekeywords={true, false, null},
    stringstyle=\color{red},
    keywordstyle=\color{blue}\bfseries,
    commentstyle=\color{gray},
    showstringspaces=false
}

\newcommand{\FIG}[3]{
\begin{minipage}[b]{#1cm}
\begin{center}
\includegraphics[width=#1cm]{#2}\\
{\scriptsize #3}
\end{center}
\end{minipage}
}

\newcommand{\editage}[1]{}\newcommand{\noeditage}[1]{#1}

\onecolumn

\begin{document}

\tikzstyle{startstop} = [rectangle, rounded corners, minimum width=3cm, minimum height=1cm,text centered, draw=black, fill=red!30]
\tikzstyle{process} = [rectangle, minimum width=3cm, minimum height=1cm, text centered, draw=black, fill=blue!20]
\tikzstyle{decision} = [diamond, minimum width=3cm, minimum height=1cm, text centered, draw=black, fill=green!30]
\tikzstyle{arrow} = [thick,->,>=stealth]

\newcommand{\figD}{
\begin{figure*}
    \begin{tikzpicture}[node distance=2cm]

        \node (start) [process] {Robot};
        \node (sensor) [process, below of=start] {Visual Sensor};
        \node (mapping) [process, below of=sensor] {Map Construction};
        \node (update) [process, right of=mapping, xshift=5cm] {Map Update};
        \node (query) [process, below of=mapping] {User Query};
        \node (knowledge) [process, below of=query] {Knowledge Base};
        \node (ogn) [process, right of=knowledge, xshift=5cm] {Object Goal Navigation (OGN)};
        \node (mrid) [process, below of=ogn] {MRID};
        \node (mapless) [process, right of=mrid, xshift=5cm] {Mapless OGN};

        \draw [arrow] (start) -- (sensor);
        \draw [arrow] (sensor) -- (mapping);
        \draw [arrow] (mapping) -- (query);
        \draw [arrow] (mapping) -- (update);
        \draw [arrow] (query) -- (knowledge);
        \draw [arrow] (knowledge) -- (ogn);
        \draw [arrow] (ogn) -- (mrid);
        \draw [arrow] (mrid) -- (mapless);

    \end{tikzpicture}
\caption{Algorithm flowchart.}\label{fig:D}
\end{figure*}
}

\renewcommand{\figD}{
\begin{figure*}
\hspace*{5mm}\FIG{17}{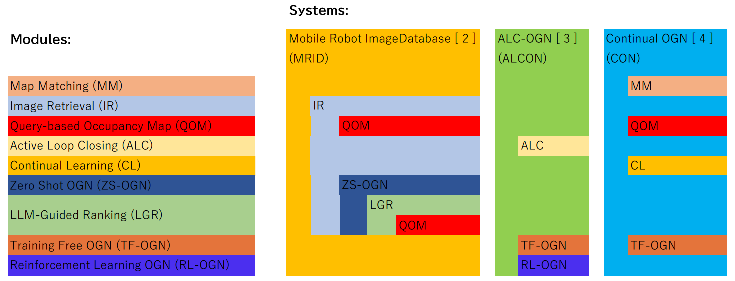}{}\vspace*{1mm}\\
\caption{The relationship between systems and modules. To avoid unnecessary complexity, common modules such as SLAM are omitted in this figure.}\label{fig:D}
\end{figure*}
}

\newcommand{\tabA}{
\begin{table*}[t]
    \centering
    \caption{Comparison of SPL results} \label{tab:A}
    \begin{tabular}{llll}
        \hline
        Method & 00800-TEEsavR23oF & 00801-HaxA7YrQdEC & 00802-wcojb4TFT35 \\
        \hline
        Random Frontier & 0.366 & 0.338 & 0.324 \\
        LGR (Ours) & 0.426 & 0.359 & 0.335 \\
        \hline
    \end{tabular}
\end{table*}
}

\newcommand{\figA}{
\begin{figure}[t]
\centering
\hspace*{5mm}
\FIG{2.3}{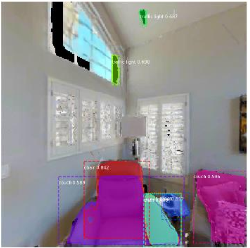}{}%
\FIG{2.3}{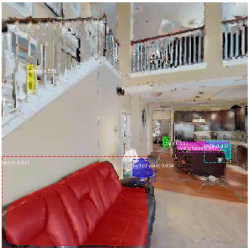}{}%
\FIG{2.3}{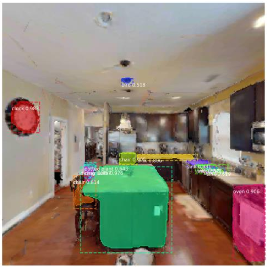}{}
n\caption{Examples of mask R-CNN outputs.}\label{fig:A}
\end{figure}
}

\newcommand{\figB}{
\begin{figure}[t]
\centering
\hspace*{5mm}\FIG{8}{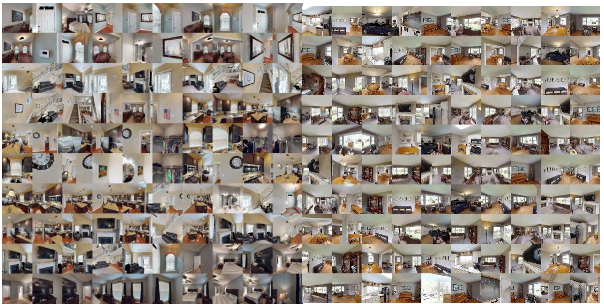}{}%
\caption{Examples of 100 target objects.}\label{fig:B}
\end{figure}
}

\newcommand{\figC}{
\begin{figure}[t]
\centering
\hspace*{5mm}\FIG{8}{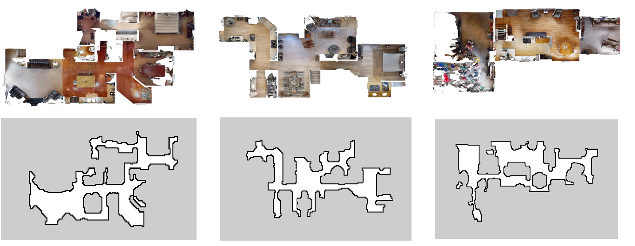}{}\vspace*{2mm}\\
\scriptsize
\hspace*{1.2cm} 800 \hspace*{2.2cm} 801 \hspace*{2cm} 802
\caption{Bird's eye views and occupancy grid maps.}\label{fig:C}
\end{figure}
}

\newcommand{\CO}[1]{}
\renewcommand{\CO}[1]{#1}
\CO{

\title{\bf\Large%
LGR: LLM-Guided Ranking of Frontiers for Object Goal Navigation
}

\author{Mitsuaki Uno, Kanji Tanaka, Daiki Iwata, Yudai Noda, Shoya Miyazaki, Kouki Terashima
\thanks{Our work has been supported in part by JSPS KAKENHI Grant-in-Aid for Scientific Research (C) 20K12008 and 23K11270.}
\thanks{The authors are with the Department of Engineering, University of Fukui, Japan.  {\tt \{ ha210233, tnkknj, mf240050, mf240327, ha211418, mf240271 \}@g.u-fukui.ac.jp}}
}

\maketitle

\begin{abstract}
Object Goal Navigation (OGN) is a fundamental task for robots and AI, with key applications such as mobile robot image databases (MRID). In particular, mapless OGN is essential in scenarios involving unknown or dynamic environments. This study aims to enhance recent modular mapless OGN systems by leveraging the commonsense reasoning capabilities of large language models (LLMs). Specifically, we address the challenge of determining the visiting order in frontier-based exploration by framing it as a frontier ranking problem. Our approach is grounded in recent findings that, while LLMs cannot determine the absolute value of a frontier, they excel at evaluating the relative value between multiple frontiers viewed within a single image using the view image as context. We dynamically manage the frontier list by adding and removing elements, using an LLM as a ranking model. The ranking results are represented as reciprocal rank vectors, which are ideal for multi-view, multi-query information fusion. We validate the effectiveness of our method through evaluations in Habitat-Sim.
\end{abstract}

\section{Introduction}

Object Goal Navigation (OGN) is a task in which a robot explores and locates a user-specified object within a workspace, widely studied in robotics and artificial intelligence \cite{OgnSurvey}. If object locations are pre-recorded on a map, the most efficient method is to retrieve the object from the mobile robot image database \cite{IROS2006kanji, arXivIwata, IROSW2024}. However, in unknown environments or when map information is unreliable, mapless OGN is essential. Existing OGN methods include end-to-end approaches, which directly generate action commands from sensor data \cite{EndToEndOGN}, but these require extensive training data and high computational costs. Recently, modular approaches have gained popularity \cite{ModularOGN}, where scene understanding, object detection, map updates, and planning are separated and optimized independently. This enhances system flexibility and efficiency while allowing focused research on specific modules. This study adopts a modular OGN approach, focusing specifically on the action planning module. 

As a baseline exploration algorithm, this study employs the Frontier-Based Exploration (FBE) method \cite{YamauchiB1997}. This FBE method directs exploration towards the boundary between known and unknown areas (frontiers). The robot, starting with limited information, must access unexplored regions to gain new knowledge. The frontier method systematically expands knowledge by navigating to these boundaries, theoretically ensuring complete exploration of accessible areas with perfect sensors and control \cite{YamauchiB1997}. However, real-world robots suffer from sensor noise and control errors, making performance hardware-dependent. As environments become more complex, selecting optimal frontiers significantly impacts exploration efficiency. Thus, improving frontier selection and visit ordering is crucial \cite{HeuristicsFrontier}. 
The use of FBE as a guide in OGN is an ongoing research topic in recent studies \cite{NEURIPS2024_098491b3, 10.1007/978-3-031-73254-6_10, 10802670, NEURIPS2024_459d93da}.

Among the unresolved issues in FBE, our study focuses on determining the optimal visiting-order of frontiers. This problem is NP-hard, making exact solutions computationally infeasible. Heuristic methods are commonly used, including Random Frontier Selection and Nearest Frontier Selection, where the robot visits the closest frontier \cite{HeuristicsFrontier} and or most promising directions recommended by high-level reasoning \cite{zhou2023esc}. However, these suboptimal methods can lead to inefficient movements, such as repeatedly traversing between distant points. State-of-the-art methods formulate exploration as the Traveling Salesman Problem (TSP) \cite{TSP} or the Orienteering Problem \cite{Orienteering}, utilizing deep reinforcement learning and constrained optimization to improve efficiency. Although these methods show promise, they often depend on specific datasets and lack flexibility in unknown environments. This highlights the need for more generalized exploration strategies.

} 

To address these challenges, we propose an LLM-based frontier ranking method, LLM-Guided Ranking (LGR), which utilizes the common-sense reasoning ability of Large Language Models (LLMs) \cite{SurveyLLM}. 
Unlike traditional machine learning-based methods that require extensive training data, LLMs demonstrate strong generalization capabilities in unseen environments. In our context, LLMs are expected to enhance object search through free-text queries and zero-shot inference \cite{LGX, NavGPT, NaviLLM2}. Specifically, our approach is grounded in recent findings that, while LLMs cannot determine the absolute value of a frontier, they excel at evaluating the relative value between multiple frontiers viewed within a single image using the view image as context. We dynamically manage the frontier list by adding and removing elements, using an LLM as a ranking model. The ranking results are represented as reciprocal rank vectors, which are ideal for multi-view, multi-query information fusion. Additionally, we design an optimized prompt generation method to structure the frontier ranking as an optimization problem, ensuring consistent decision-making even in dynamically changing environments. This enables more efficient and adaptive object search by overcoming the limitations of conventional exploration methods.
Experimental validation in Habitat-Sim environments \cite{Habitat} demonstrates the effectiveness of our approach.

\section{Related Work}

Despite being a classical exploration algorithm proposed nearly 30 years ago \cite{YamauchiB1997}, the frontier-based exploration method continues to attract increasing attention due to its desirable properties. These include guaranteed coverage, simplicity, high efficiency, flexibility in integration with other perception and planning modules, and adaptability to various environments. A wide range of studies have aimed to improve frontier-based exploration. For example, approaches incorporating cost functions to enhance exploration efficiency have been proposed \cite{CostExploration}. Some studies introduce frontier selection algorithms that consider not only travel distance minimization but also information gain \cite{InformationExploration}, demonstrating improved exploration performance. Additionally, deep learning-based approaches have been explored, where an exploration strategy is learned to adapt to specific environments, often outperforming conventional methods \cite{ActiveNeuralSLAM}. In multi-agent exploration, cooperative frontier allocation among multiple robots has been studied to maximize efficiency \cite{MutltiAgentExploration}. However, many of these methods rely on training data or specific environmental characteristics, limiting their general applicability. Given this background, our study aims to leverage the commonsense reasoning capability of large language models (LLMs) to enhance the flexibility of exploration strategies.

Recently, LLM-based approaches have revolutionized various real-world applications, including ChatGPT \cite{SurveyLLM}. Specifically, LLMs have demonstrated remarkable capabilities in natural language processing \cite{NLP4LLM}, code generation \cite{CodeGenerationLLM}, and interactive agents \cite{InteractiveAgentLLM}, among other fields. Moreover, their potential applications in robotic navigation are being actively explored \cite{OgnSurvey}, with the expectation of improving decision-making through reasoning. For instance, the LGX method \cite{LGX} formulates navigation as a two-stage decision-making problem: (1) selecting a room and (2) exploring within the selected room. By integrating state-of-the-art open-vocabulary object detection \cite{OpenVoc} and grounding techniques \cite{Grounding}, LGX has demonstrated high success rates in object-goal navigation (OGN). Furthermore, some studies leverage textual descriptions of the environment to infer structural information and obstacle locations, while others focus on understanding human instructions for navigation. However, their applicability is limited in real-world exploration tasks. Challenges such as adapting to dynamic environments, ensuring real-time performance, and improving the reliability of LLM-based reasoning remain unresolved. Therefore, there is a growing need for exploration strategies that integrate LLMs to enhance adaptability in unknown environments.

This paper focuses on the "view-to-prompt" approach, which translates individual views (rather than a map) to LLM prompts and explores the potential of LLMs as a tool for ranking frontiers.
A key advantage of the "view-to-prompt" approach is its fixed prompt length---matching a single view image---ensuring effectiveness for the LLM, as verified by multiple studies. Other benefits include consistency with the scene's field of view and occlusions, along with reduced storage and transmission overhead.
While demonstrated in recent OGN-related studies~\cite{LGX, NavGPT, NaviLLM2, NEURIPS2024_098491b3, 10.1007/978-3-031-73254-6_10, 10802670, NEURIPS2024_459d93da}, the "view-to-prompt" approach remains underexplored in the context of frontier ranking. Our approach could extend to a "map-to-prompt" strategy, where entire maps are translated to LLM prompts. For example, \cite{10342512} recently explored constructing a frontier map and generating prompts based on frontier cell semantics. 
This work closely aligns with ours, as both ultimately produce ranked frontiers. However, as a robot explores, accumulated multi-view map knowledge may lead to prompts that are too complex for current LLMs. Additionally, methods for encoding navigation costs into LLM prompts remain unexplored.
While advanced LLMs may enable "map-to-prompt" in the future, this paper focuses on "view-to-prompt".

This paper focuses on the "view-to-prompt" approach, which translates individual views (rather than a map) to LLM prompts and explores the potential of LLMs as a tool for ranking frontiers.
A key advantage 
of the "view-to-prompt" approach
is its fixed prompt length---matching a single view image---ensuring simplicity for the LLM, as verified by multiple studies. Other benefits include consistency with the scene's field of view and occlusions, along with reduced storage and transmission overhead.
While demonstrated in recent OGN-related studies \cite{LGX, NavGPT, NaviLLM2, NEURIPS2024_098491b3, 10.1007/978-3-031-73254-6_10, 10802670, NEURIPS2024_459d93da}, the "view-to-prompt" approach remains underexplored in the context of frontier-ranking. 
Our approach could extend to a "map-to-prompt" strategy, where entire maps are translated to LLM prompts. For example, \cite{10342512} recently explored constructing a frontier map and generating prompts based on frontier cell semantics. This work closely aligns with ours, as both ultimately produce ranked frontiers. However, as a robot explores, accumulated multi-view map knowledge may lead to prompts too complex for current LLMs. Additionally, methods for encoding navigation costs into LLM prompts remain unexplored.
While advanced LLMs may enable "map-to-prompt" in the future, this paper focuses on "view-to-prompt."

Our study is closely related to our previous works \cite{IROS2006kanji, IROSW2024, arXivIwata}. The mobile robot image database (MRID) in \cite{IROS2006kanji} and its follow-up studies structured a robot's visual experiences as viewpoint-associated images, enabling similarity-based image retrieval. These retrieved images were then summarized into a query-based occupancy map (QOM), which served as prior knowledge for further exploration. However, MRID assumed a static database, with no demonstration of adding or removing items.
More recently, the continual object goal navigation (CON) in \cite{IROSW2024} addressed multi-robot cooperative OGN, where robots dynamically updated the database through OGN tasks. In this multi-robot continual learning scenario, QOM was utilized as a lightweight yet rich knowledge representation that could be transferred between robots. However, the relevance between the query object and the observed object was naively evaluated based on appearance similarity, and the use of LLM was not implemented.
Furthermore, the active loop-closing object goal navigation (ALCON) in \cite{arXivIwata} 
pointed out that 
active loop closing (ALC) \cite{ALCorg}
(an essential task in active SLAM) 
can be reformulated as Object Goal Navigation (OGN) in cases with long travel distances.
ALCON improved long-distance ALC's efficiency by predicting revisit points based on semantic frontier selection. This result suggested that mapping not only supports OGN tasks but also benefits from OGN itself.
These prior works share a common focus on evaluating frontier relevance based on semantic information. In contrast, our study introduces a novel approach that is orthogonal yet complementary to these by leveraging commonsense reasoning.

\noeditage{
\figD
}

\section{System Configuration}

In this section, we introduce the mobile robot system we are developing and position this study within its framework. The system structure is illustrated in Figure \ref{fig:D}.

The mobile robot is equipped with a visual sensor and a mobility mechanism. This system constructs a map based on the robot's visual experience and updates it according to environmental changes, enabling accurate self-localization and adaptive navigation. 
The robot acquires surrounding information using a visual sensor and generates an environmental map from the collected data. 
During this process, the system detects environmental changes by comparing the visual information with the existing map and updates the map accordingly. Additionally, a user interface provides a query function for the map knowledge base, allowing users to request specific information. This knowledge base leverages the robot's accumulated visual data, contributing to improved navigation and task execution accuracy.

\subsection{Mobile Robot Image Database (MRID)}

In this study, we utilize the Mobile Robot Image Database (MRID) to achieve Object Goal Navigation (OGN). MRID takes user-specified object descriptions as input and outputs a query-based occupancy map (QOM) indicating the regions where the specified object is likely to be found. Furthermore, MRID includes a knowledge transfer function that allows the object occupancy map to be shared with other robots, enabling continual object goal navigation (CON).

The OGN system targeted in this study leverages MRID for object exploration and navigation. First, the user queries MRID for information about the specified object. If MRID can determine the object's location, a map-based OGN process is executed. If the object is found, the task is completed. However, if the object is not found, the system transitions to a mapless OGN approach to continue the search. On the other hand, if MRID cannot determine the object's location from the beginning or if any failure occurs in the mapping system, the system directly executes mapless OGN for object exploration. Thus, the triggering of the mapless OGN module is determined by MRID. In this paper, we focus on the pure mapless OGN problem after the trigger event.

\section{LGR Approach}

The method formulates the order of frontier visits as a ranking problem and enables the robot to query the LLM to facilitate more efficient exploration. 

First, the robot identifies candidate frontiers at the boundary between known and unknown regions and lists them. Then, the robot queries the LLM with a prompt such as:
\begin{quote}
"The target object is a 'red chair.' Among the following frontiers, which should be prioritized for exploration?"
\end{quote}

The LLM considers the semantic information of the environment and the typical spatial arrangement of objects to recommend an exploration order. Finally, the robot follows this ranking to navigate efficiently toward the target object.

Specifically, when the robot explores the environment, it defines a set of frontier candidates located at the boundary between known and unknown regions as follows:
\[
\mathcal{F} = \{ f_1, f_2, \dots, f_N \}
\]
Each frontier \( f_i \) is represented by its coordinates \( \mathbf{x}_i \in \mathbb{R}^2 \) (or \( \mathbb{R}^3 \)).

The query sent to the LLM is structured as follows:
\begin{quote}
"The target object is a 'red chair.' Among the following frontier candidates, which should be prioritized for exploration?"
\end{quote}

The LLM ranks each frontier \( f_i \) according to an estimated priority score \( r_i \), where \( r_i \) is a natural number, and smaller values indicate higher priority.

Additionally, following the methodology in \cite{DragonOGN}, the robot may evaluate the distance from its current position \( \mathbf{x}_r \) to each frontier as:
\[
d(f_i) = | \mathbf{x}_r - \mathbf{x}_i |_2
\]
A weighting function \( w_i \), which is a monotonically decreasing function of this distance, can also be considered.

By integrating these weights and the reciprocals of the ranking values, a score vector representing the priority of each frontier is obtained as follows:
\[
S = [ w_1\frac{1}{r_1}, \cdots, w_N\frac{1}{r_N} ]
\]
Given this score vector \( S \), the optimal frontier is determined by:
\[
i^* = \arg\max_{i \in [1, N]} S[i]
\]
The robot then moves to this frontier, acquires new environmental information, and continues the exploration process.

\subsection{Reciprocal Rank Feature (RRF)}

One of the remarkable characteristics of the reciprocal rank feature (RRF) vector is its ability to fuse information from different observations \cite{GoogleRRF}. When we treat the response from the LLM to a prompt as an observation, the results of that observation can be expressed in the form of an RRF vector. From a new viewpoint, a newly generated prompt based on an updated list of frontiers leads to a relative ranking determined by the LLM. This ranking result is also represented in the RRF format.

Now, assuming that the relative rankings generated by the LLM are considered independent observations, their results can be fused by simply adding the corresponding RRF vectors. Of course, as the robot moves to a new viewpoint, it must update the frontier list $\mathcal{F}$ by removing unnecessary frontiers and adding newly discovered (potentially valuable) frontiers. Optimally designing this frontier list update rule is beyond the scope of this paper and remains an ongoing research topic. However, a prototype implementation of one such approach is introduced in Section~\ref{sec:prompter}.

Frontier cells are compactly represented as scalar values in the form of reverse rank scores. This contrasts with typical feature maps that have vector or tensor values as cell values (e.g., appearance, semantics, spatial). The frontier is expressed as relative rank values through LLM-Guided Ranking, which enables our score map representation. To reflect the idea of constructing a map based on queries to the LLM, we will refer to this as a query-based occupancy map (QOM).

Unlike existing methods such as LGX \cite{LGX}, which assume a top-down workspace model (e.g., room-object relationships), we adopt a non-top-down workspace model that more closely reflects real-world environments (e.g., room-object-room). To address this problem, we propose a frontier-driven exploration approach in which frontier selection is determined using an LLM. This enables zero-shot object goal navigation using a generalized frontier method, making exploration more adaptable to diverse environments.

Among the key components of our system, the image recognition module (Section~\ref{sec:recognition}) and the motion planning module (Section~\ref{sec:planner}) follow well-established best practices. On the other hand, prompt design, which governs interactions between the robot and the LLM, constitutes a major contribution of this paper and is discussed in detail in Section~\ref{sec:prompter}.

\noeditage{
\figA
}

\subsection{Image Recognition Module}\label{sec:recognition}

The image recognition module employs Mask R-CNN \cite{MaskRCNN} to enable the robot to recognize objects in its surroundings (Fig. \ref{fig:A}). 
The robot captures images in eight directions (front, back, left, right, and four diagonal directions) to recognize objects in its surroundings (Fig. \ref{fig:A}) and updates an occupancy grid map based on its observations. 
These images are then processed by Mask R-CNN to detect objects within each frame. The detected objects, along with their class labels and positional information, are structured as text data and forwarded to the motion planning module.

In this study, we use a closed-vocabulary Mask R-CNN; however, future extensions to open-vocabulary recognition \cite{OpenVoc} can be readily implemented.

\subsection{Motion Planning Module}\label{sec:planner}

For decision-making, the robot utilizes an occupancy grid map (OGM) \cite{OGM} and the frontier-based exploration method. 
The OGM represents the environment by classifying each grid cell into three states: unknown, occupied, or free. The frontier method identifies the boundaries between known (occupied or free) and unknown regions to facilitate efficient exploration. Additionally, the LLM estimates the likelihood of target object presence and prioritizes frontiers as subgoals.

The robot determines its movement using the A* algorithm, computing the shortest path to its selected sub-goal. After reaching the destination, it updates the environmental map and continues exploration. The robot captures images in eight directions, processes them with Mask R-CNN, and uses the LLM to estimate object presence probabilities, which guide its exploration direction. The system ranks frontiers in terms of priority and selects the highest-ranked one as the next sub-goal. The A* algorithm then plans the path, and the robot executes the movement. Upon arrival, the environmental map is updated, and object detection is re-executed to refine the exploration process.

Through this iterative process, the robot autonomously integrates object recognition, environmental understanding, and exploration planning to achieve intelligent navigation. In the current implementation, an idealized bumper sensor is used to prevent robot failure due to collisions with obstacles.

\section{Prompt Design}\label{sec:prompter}

In this study, we generate prompt texts to be sent to a large language model (LLM) based on the information obtained from the object detection model. Proper prompt design is crucial as it enhances environmental understanding and improves exploration efficiency, requiring careful formulation.

Each time the robot moves to a new viewpoint, it performs a geodesic rotation. Specifically, the robot rotates at 45-degree increments from 0 to 360 degrees, capturing images in eight different orientations. While setting the number of images to eight is not mandatory, we follow the configuration used in \cite{LGX}. Under this setting, assuming the camera's field of view is 45 degrees, it should be noted that the full 360-degree view can be covered.

To illustrate the prompt design methodology, we use an example of object recognition results, as shown in Figure~\ref{fig:A}.

First, considering the spatial features of the environment, we utilize the results of object detection to construct prompts for the LLM. In this study, we design the prompt shown in Figure~\ref{fig:prompt_category} to classify the environment into categories.

\begin{figure*}[h]
    \centering
    \begin{lstlisting}[language=json, basicstyle=\ttfamily\footnotesize]
    {"The detected objects are: {od}. Which room category from {rlc} is most likely?"}
    \end{lstlisting}
    \caption{Prompt for room category classification}
    \label{fig:prompt_category}
\end{figure*}

Here, \texttt{{od}} represents the list of detected objects, and \texttt{{rlc}} denotes the candidate room categories. In this study, we use the list shown in Figure~\ref{fig:room_categories} for \texttt{{rlc}}.

\begin{figure*}[h]
    \centering
    \begin{lstlisting}[language=json, basicstyle=\ttfamily\footnotesize]
    ["bathroom", "bedroom", "reception room", "laundry room", "kitchen",
     "home office", "living room", "wall"]
    \end{lstlisting}
    \caption{List of room categories}
    \label{fig:room_categories}
\end{figure*}

Based on the results obtained from the above prompt, we perform ranking using the prompt shown in Figure~\ref{fig:ranking_prompt}.

\begin{figure*}[h]
    \centering
    \begin{lstlisting}[language=json, basicstyle=\ttfamily\footnotesize]
    Now, could you please list all the selected locations from Step 1 in a concise manner,
    one location per line, like this:
    Step 1: [location]
    Step 2: [location]
    ...
    Please ensure the response is limited to only one concise list and does not contain
    any additional explanations or variations.

    Now, I am trying to determine the most likely place where a "{og}" (defined in step 11)
    might be found. Based on the selected locations from the previous steps,
    please identify the most likely location(s).

    ### Important Rules:
    1. Even if a location appears multiple times in the steps, you must consider each 
       occurrence individually and assign it a unique rank.
    2. Do not group or combine multiple occurrences of the same location.
    3. All {len(responses)} steps must be ranked from 1 to {len(responses)} without 
       omitting any step.

    ### Expected Output:
    Please provide the ranked locations in the following format:
    1. [Location from Step X]
    2. [Location from Step Y]
    3. [Location from Step Z]
    ...
    {len(responses)}. [Location from Step W]

    Ensure that the ranking explicitly includes the step number along with the location
    (e.g., "living room from Step 2").
    \end{lstlisting}
    \caption{Prompt for ranking locations}
    \label{fig:ranking_prompt}
\end{figure*}

By using this prompt, we obtain results with multiple {\it steps} as shown in Figures~\ref{fig:step11_output}, \ref{fig:step12_output}, and \ref{fig:step13_output}. Specifically, in Step 11, we determine which location corresponds to each of the eight directional images. In Step 12, we identify the most likely room where the target object is located. Finally, in Step 13, we rank the locations based on the likelihood of finding the target object.

\begin{figure*}[h]
    \centering
    \begin{lstlisting}[basicstyle=\ttfamily\footnotesize]
    Response for od1: kitchen
    Response for od2: bedroom
    Response for od3: kitchen
    Response for od4: bathroom
    Response for od5: bedroom
    Response for od6: kitchen
    Response for od7: living-room
    Response for od8: living-room
    \end{lstlisting}
    \caption{Detection results for each object}
    \label{fig:output_response}
\end{figure*}

\begin{figure}[h]
    \centering
    \begin{lstlisting}[basicstyle=\ttfamily\footnotesize]
    Step 1: kitchen
    Step 2: bedroom
    Step 3: kitchen
    Step 4: bathroom
    Step 5: bedroom
    Step 6: kitchen
    Step 7: living-room
    Step 8: living-room
    \end{lstlisting}
    \caption{Output of Step 11}
    \label{fig:step11_output}
    
    \centering
    \begin{lstlisting}[basicstyle=\ttfamily\footnotesize]
    living-room
    \end{lstlisting}
    \caption{Output of Step 12}
    \label{fig:step12_output}
    
    \centering
    \begin{lstlisting}[basicstyle=\ttfamily\footnotesize]
    1. living-room from Step 7
    2. living-room from Step 8
    3. bedroom from Step 2
    4. bedroom from Step 5
    5. kitchen from Step 1
    6. kitchen from Step 3
    7. kitchen from Step 6
    8. bathroom from Step 4
    \end{lstlisting}
    \caption{Output of Step 13}
    \label{fig:step13_output}
\end{figure}

\subsection{Determining Orientation}

The robot's direction of movement is determined based on the ranking results obtained by sending the prompt texts from Figures~\ref{fig:prompt_category}, \ref{fig:room_categories}, and \ref{fig:ranking_prompt} to the LLM. In this method, the highest-ranked orientation (rank 1) in the LLM's output is selected as the next movement direction.

Based on the selected orientation, the robot's movement is controlled to improve exploration efficiency. Additionally, the remaining seven directions that were not selected in the ranking are still utilized for accumulating environmental information. Specifically, among the unselected orientations, those containing unexplored regions (frontiers) are retained so that their image information can be referenced in future decisions. By integrating past and newly acquired information in this manner, more optimal route selection can be achieved.

\subsection{Determining Subgoals}

The subgoal is determined from the frontiers within the movement direction selected by the LLM.

\subsection{Selecting Movement Direction}

First, the highest-ranked direction from the LLM ranking results is obtained. A frontier (unexplored region) within that direction is randomly selected and set as the subgoal.

\subsection{Moving Toward the Subgoal}

Once a frontier cell is selected as the next subgoal, the robot begins moving toward that frontier. This movement may be blocked by obstacles, which are detected by the bumper sensor. In such cases, the robot stops, updates the OGM, and reselects the next subgoal based on the updated OGM. During these planning and replanning steps, not only the newly observed frontiers but also previously observed and scored frontiers can be reused.
If the length of the frontier list does not meet the preset threshold, all frontiers in the list, both new and old, could potentially be ranked in a single LLM query, which would be the most reliable. However, this method quickly loses reliability as the frontier list grows longer. We have observed that current LLMs cannot adequately compare a very long frontier list. In such cases, using the aforementioned "view-to-prompt" approach helps keep the prompt length fixed. By repeating this process, the robot continuously updates its exploration path and incorporates newly gathered information to determine the optimal route.

\section{Experiment}

The objective of this experiment, as described in Chapter 4, is to leverage the strength of the frontier method in generalization performance to achieve zero-shot object-goal navigation, enabling exploration in more general environments. 
For proof-of-concept validation, we compare the performance of the proposed LGR method with its ablation, the random frontier method.

\subsection{Dataset (Habitat Workspace)}

In this study, we used the Habitat Matterport 3D Research Dataset (HM3D) MINIVAL dataset\cite{Habitat} within the Habitat environment to conduct object-goal navigation. This dataset simulates realistic 3D indoor spaces and provides diverse information on object locations and spatial structures.

In this experiment, only the first floor of each workspace was designated as the exploration area, and the experiments were conducted in a simulation environment.

\noeditage{
\figB
}

\subsection{Experimental Setup}

In this experiment, 100 target objects were set in each workspace (e.g., Fig. \ref{fig:B}), and exploration was initiated from different starting points.
The target objects and starting points were randomly selected and paired while ensuring a certain distance between them.
The dataset composed of these pairs was shared between the proposed method and the baseline method to ensure a fair comparison.

The following technologies were used in the experiment:
\begin{itemize}
\item Large Language Model (LLM): Google Gemini
\item Object Detection Model: Mask R-CNN
\item Path Planning Algorithm: A* Method
\end{itemize}

\subsection{Experimental Procedure}
The robot scans the surrounding environment using an omnidirectional pre-casting method, collecting image data in eight directions (front, back, left, right, and four diagonal directions).

\begin{enumerate}
    \item The robot performs object detection based on the acquired images from eight directions and identifies the locations of recognized objects.
    \item Based on ranking results from the LLM, the robot determines the prioritized locations for reaching the target object.
    \item The robot detects frontiers (boundaries between known and unknown areas) in each direction.
    \item Based on LLM ranking, a frontier within the highest-ranked direction is randomly selected and set as the next subgoal.
    \item The robot moves using the shortest path search based on the A* method.
    \item After moving, the robot acquires new images from eight directions and updates the map.
    \item The exploration process is repeated until the target object is detected.
\end{enumerate}

\subsection{Performance Metric}
In this study, SPL (Success Weighted by Path Length) \cite{OgnSurvey} was adopted as the evaluation metric, as it simultaneously evaluates both movement efficiency and success rate.

SPL is defined by the following equation:
\begin{equation}
SPL = \frac{1}{N} \sum_{i=1}^{N} S_i \cdot \frac{l_i^*}{\max(l_i, l_i^*)}
\end{equation}

where:
\begin{itemize}
    \item \( N \): Total number of navigation sessions
    \item \( S_i \): 1 if the session is successful, 0 if unsuccessful
    \item \( l_i^* \): Shortest path length to reach the target
    \item \( l_i \): Actual path length traveled by the robot
\end{itemize}

Since SPL reflects both movement efficiency and success rate, it is an appropriate metric for evaluating object-goal navigation performance.

\noeditage{
\tabA
}

\noeditage{
\figC
}

The results of the experiment are shown in Table~\ref{tab:A}.
The bird's-eye view of the workspace and the occupancy grid map for the workspaces are shown in Figure~\ref{fig:C}.
The proposed method was compared with the random frontier method in three workspaces: 
00800-TEEsavR23oF 1F (800), 00801-HaxA7YrQdEC 1F (801), 00802-wcojb4TFT35 2F (802), and their respective SPL values are presented.
The results indicate that the proposed method outperformed the conventional method in all workspaces.
For the workspace "800", the SPL of the proposed method was 0.426, outperforming the random frontier method's 0.366. 
This result indicates that the proposed method achieved more efficient path exploration, reducing travel distance while improving success rates.
For the workspace "801",
the SPL of the proposed method was 0.359, surpassing the random frontier method's 0.328. This result suggests that the proposed method conducted more efficient exploration compared to the random frontier method.
For the workspace "802",
the SPL of the proposed method was 0.335, exceeding the random frontier method's 0.324. This result confirms that the proposed method conducted more efficient exploration compared to the random frontier method.
Notably, in the workspace "800", SPL improved from 0.366 to 0.426, a 16.4\% increase. Additionally, improvements of 6.2\% in "801" and 3.4\% in "802" were observed, demonstrating the effectiveness of the proposed method.
The results of this experiment are sufficient as a proof-of-concept for the proposed method. More refined results will be published in the next version of this paper.

\bibliographystyle{unsrt}
\bibliography{reference}

\editage{
\figD
\figA
\figB
\tabA
\figC
}

\end{document}